\documentclass{article}

\PassOptionsToPackage{numbers,square}{natbib}

\usepackage[final]{nips_2017}

\usepackage[utf8]{inputenc} 
\usepackage[T1]{fontenc}    
\usepackage{courier}       
\usepackage{hyperref}       
\usepackage{url}            
\usepackage{booktabs}       
\usepackage{amsfonts}       
\usepackage{amssymb}
\usepackage{amsthm}
\usepackage{nicefrac}       
\usepackage{microtype}      
\usepackage{makecell}      %
\usepackage{graphicx}
\usepackage{subfig}
\graphicspath{ {images/} }
\usepackage[font=small,labelfont=bf]{caption}
\usepackage[shortlabels]{enumitem}
\newcommand*\samethanks[1][\value{footnote}]{\footnotemark[#1]}

\title{Real-time Egocentric Gesture Recognition on\\
       Mobile Head Mounted Displays}

\author{
  Rohit Pandey\thanks{Equal Contribution}\qquad
	Marie White\samethanks\qquad
	Pavel Pidlypenskyi\\
	\textbf{Xue Wang}\qquad
	\textbf{Christine Kaeser-Chen}\\
	\\
  Google Inc.\\
  \texttt{\{rohitpandey,mariewhite,podlipensky} \\
  \texttt{xuew,christinech\}@google.com} \\
}

\begin{document}

\maketitle

\begin{abstract}
  Mobile virtual reality (VR) head mounted displays (HMD) have become popular
  among consumers in recent years.  In this work, we demonstrate real-time
  egocentric hand gesture detection and localization on mobile HMDs. Our main
  contributions are: 1) A novel mixed-reality data collection tool to automatic
  annotate bounding boxes and gesture labels;  2) The largest-to-date egocentric
  hand gesture and bounding box dataset with more than 400,000 annotated frames;
  3) A neural network that runs real time on modern mobile CPUs, and achieves
  higher than 76\% precision on gesture recognition across 8 classes.
\end{abstract}

\section{Introduction}

Mobile virtual reality (VR) head mounted displays (HMD), such as Daydream and
GearVR, have made VR more accessible. Making users believe that they can
interact with the virtual environment is critical to immersion. Since people
interact with the real environment mostly with hands, we study how to bring hand
presence to VR. In this work, we focus on hand gesture detection and
localization. Our goal is to reliably recognize and localize hand gestures in
real-time on mobile HMD systems.

There are two main challenges to this problem:
\begin{enumerate}[nosep]
  \item There is limited dataset available on egocentric hand gestures and
    bounding boxes.  
  \item It is challenging for high-capacity machine learning
    models to run at interaction framerate on mobile devices
\end{enumerate}
To the first challenge, we propose to utilize mobile mixed
reality headset as a tool to collect data and automatically label bounding box.
With this method, we collected a large dataset of 33 people in 30 different
scenes, with a total of 406,581 annotated frames. To download the dataset,
please visit \url{https://sites.google.com/view/hmd-gesture-dataset}.

To the second challenge, we trained a neural network based on the TensorFlow
Object Detection API \cite{tf-object}. The network uses MobileNet
\cite{mobilenet} as the feature extractor and SSD head \cite{SSD} to generate
multibox predictions. When running on mobile, one forward pass
of our model takes 31.85 milliseconds on one core mobile CPU, achieving real-time
performance. 

\section{Background}
There has been extensive research on hand gesture recognition
system. \cite{Pavlovic_survey}, \cite{Mitra_survey}, \cite{Rautaray_survey} provide
execellent surveys on vision-based gesture recognition systems.

Convolutional neural networks (CNN) such as in
\cite{Molchanov_2015_CVPR_Workshops} and recurrent neural networks (RNN) such as
in \cite{Cui_2017_CVPR}\cite{Cao_2017_ICCV} have further pushed the boundary on
gesture recognition results. Unfortunately none of them has shown real-time
performance on mobile devices.

Building large and diverse datasets for
hand gestures remains challenging. Existing gestures datasets such as those from
\cite{car-app} have less than 10,000 annotated frames. The closest datasets to
our work are EgoFingers \cite{egofinger} where 93,729 frames are labeled, and EgoGesture
\cite{Cao_2017_ICCV} where 3 million frames have gesture labels but no bounding
boxes. Compare to these datasets, our dataset contains 406,581 frames with both
gesture labels and bounding boxes.

\section{Dataset}
\label{dataset}

\subsection{\textit{Label As You Go}}
In order to scale data collection, we utilize mobile mixed
reality headsets to collect automatically labeled data. 
Images in our dataset are labeled by the subjects as the images are
recorded, instead of by annotators after the fact.

We
used a Daydream View, a Google Pixel XL smartphone, and a monochrome USB camera
that is connected to
the phone and faces the world. In the headset, users can see digital video passthrough of the outward facing
camera, and hence see the real world while in VR. 

On top of video passthrough, we overlay a bounding box target on each frame in
camera image space. For
each gesture class, subjects are instructed to pose the requested gesture, and
fit their hands tightly into the rendered bounding box target. With the mixed
reality setting, this task is simply done 
through natural hand-eye coordination.
 
We vary the location of the bounding box to increase coverage of the
dataset. To further reduce time, we animate the bounding box target in a
pre-defined zigzagging trajectory which sweeps across the whole frame. As the
trajectory is predictable and easy to remember, the subjects are able to follow
the box, even as it moves to a new location. Each subject participating in data
collection was asked to pose 4
gestures on each hand: $\texttt{Thumbs\_Press}$, $\texttt{Thumbs\_Up}$,
$\texttt{Thumbs\_Down}$, and $\texttt{Peace}$. For each gesture, we instrument 3
sequences of trajectories. The bounding box
size stays the same for a single sequence, but varies from sequence to sequence.
We provide a clicker to the subjects to signal the start of each data collection
sequence. 

\begin{table}[t]
  \centering
  \begin{tabular}{lllll}
    \toprule
    & \makecell[c]{$\texttt{Thumbs\_Press}$} & \makecell[c]{$\texttt{Thumbs\_Up}$} &
    \makecell[c]{$\texttt{Thumbs\_Down}$} & \makecell[c]{$\texttt{Peace}$} \\
    \midrule
    & \includegraphics[width=2.4cm,height=1.8cm]{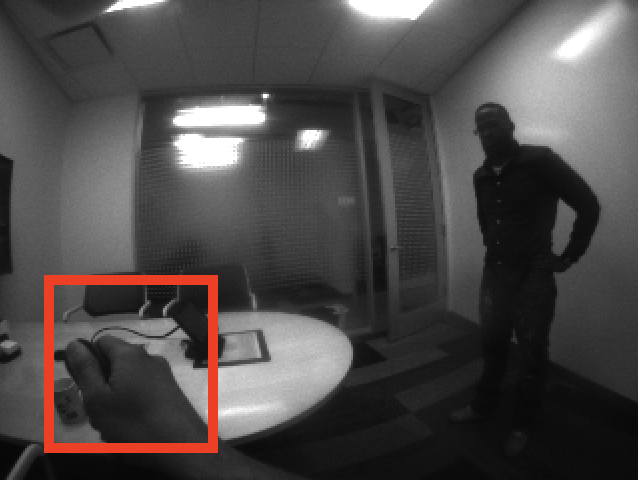}  & 
    \includegraphics[width=2.4cm,height=1.8cm]{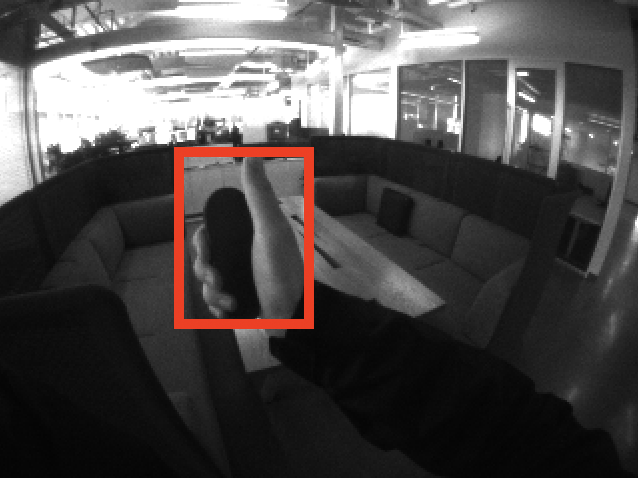} &
    \includegraphics[width=2.4cm,height=1.8cm]{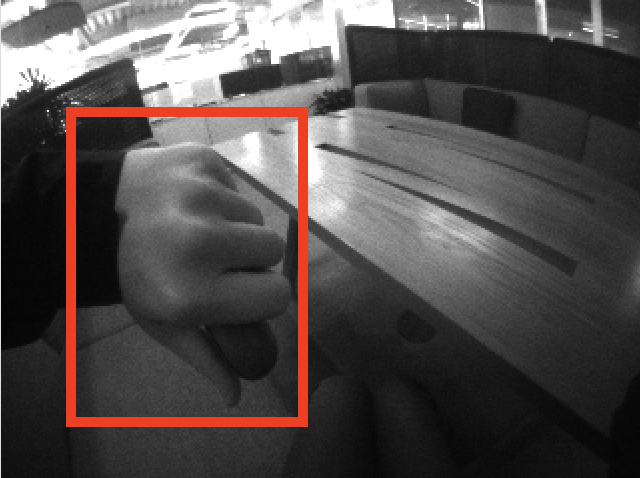} &
    \includegraphics[width=2.4cm,height=1.8cm]{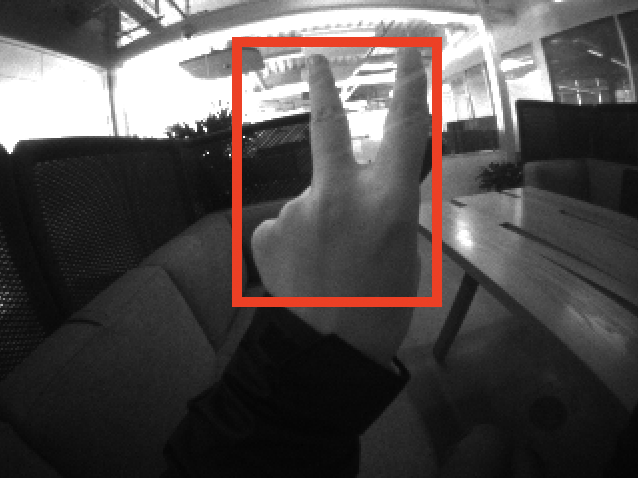} \\ 
    \midrule
    \makecell{\# of Frames \\ (\% of total)}  & 
    \makecell{113206 \\ (27.8\%)} & 
    \makecell{120716 \\ (29.7\%)} &  
    \makecell{55844 \\ (13.7\%)} & 
    \makecell{116815 \\ (18.7\%)}  \\ 
    \bottomrule
  \end{tabular}
  \vspace{-5pt}
  \caption{Sample images and gesture distribution in our dataset}
  \vspace{-18pt}
  \label{gesture-table}
\end{table}

\begin{figure}[t]
    \vspace{-10pt}
    \centering
    \subfloat[]{\includegraphics[height=3cm]{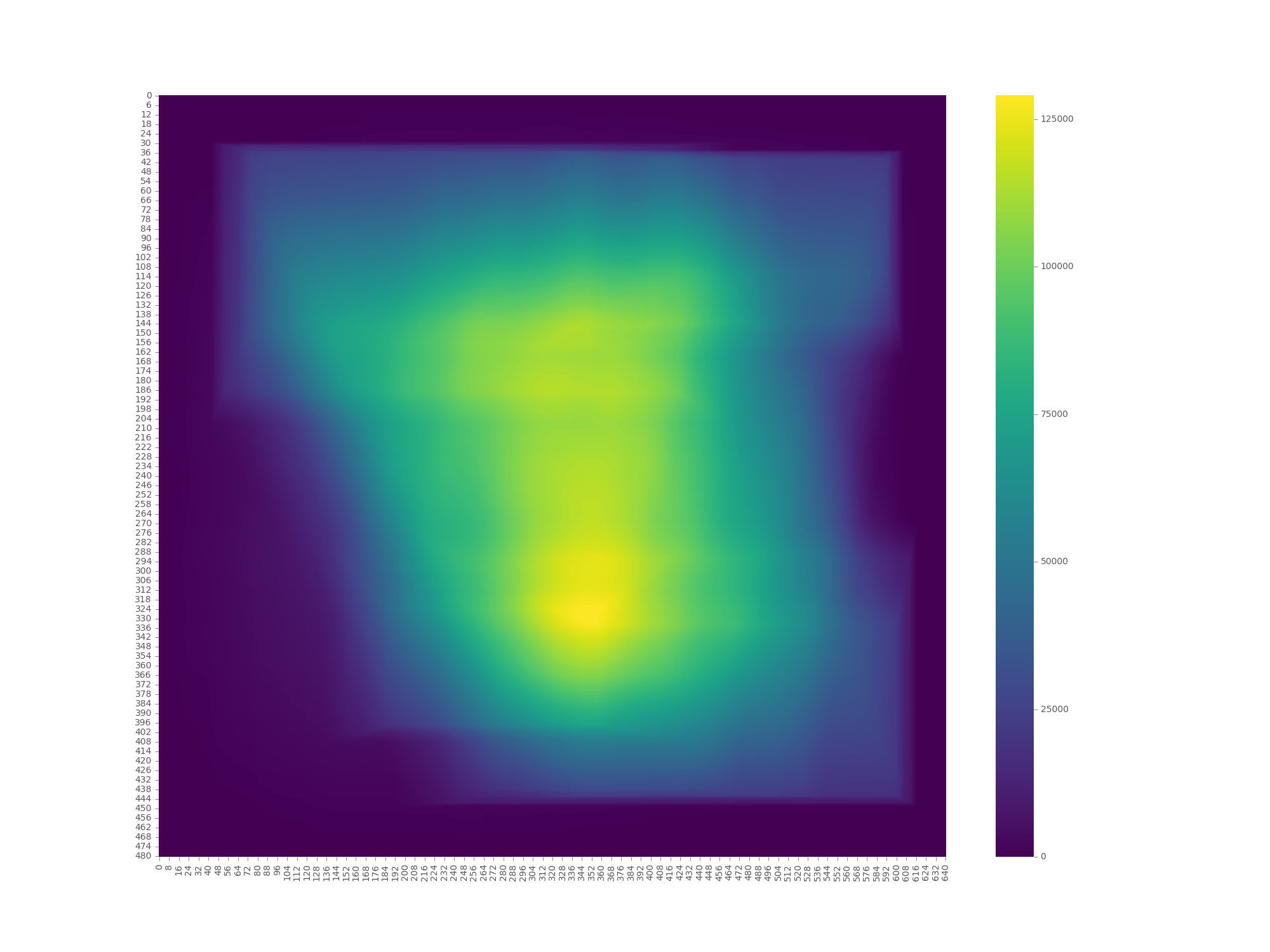}
        \label{fig:a}}\hspace*{-2em}
    \subfloat[]{\includegraphics[height=3cm]{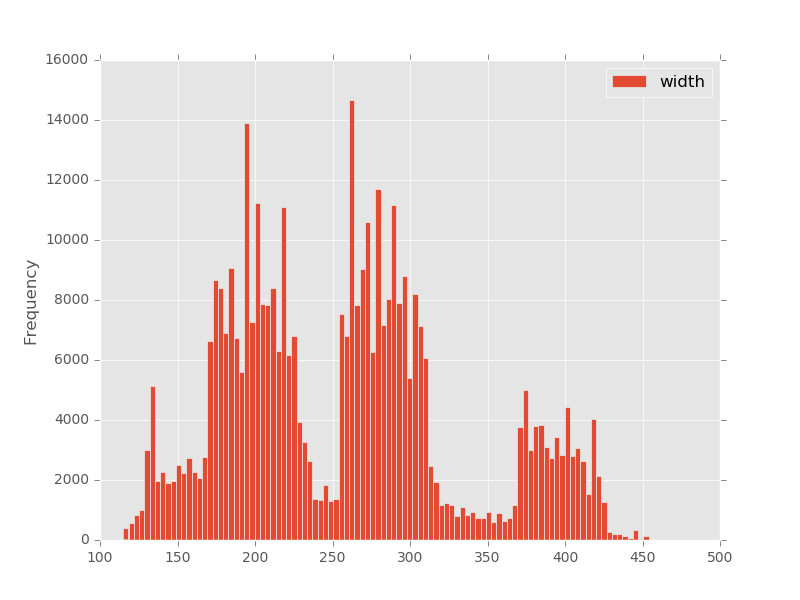}
        \label{fig:b}}\hspace*{-2em}
    \subfloat[]{\includegraphics[height=3cm]{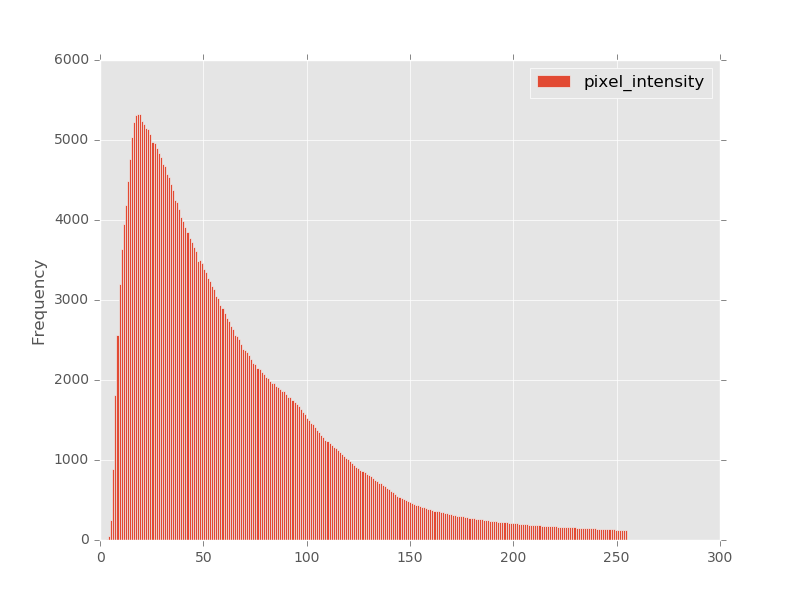}
        \label{fig:c}}\hspace*{-2em} 
    \subfloat[]{\includegraphics[height=3cm]{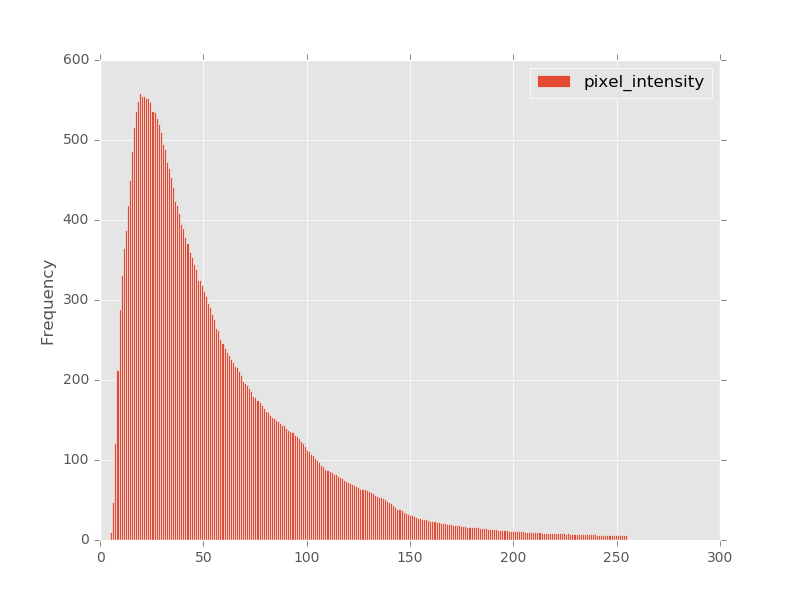}
        \label{fig:d}}\hspace*{-2em}
    \vspace{-5pt}
    \caption{Image statistics of the dataset. 
             \protect\subref{fig:a} Distribution of bounding boxes;
             \protect\subref{fig:b} Histogram of bounding box sizes; 
             \protect\subref{fig:c} Histogram of pixel intensity in all images; 
             \protect\subref{fig:d} Histogram of pixel intensity inside all
             bounding boxes; }
    \label{image-stats}
    \vspace{-15pt}
\end{figure}

\subsection{Dataset Details}
With the \textit{label as you go} approach, we built a dataset of 406,581 frames of egocentric hand gesture data. The full dataset creation process took only two days. Each frame is labeled with 
gesture class and the bounding box of hands. Our dataset contains data from 33
subjects and 4 gesture classes on each hand. Since the mixed reality setup is mobile, we
are able to collect data in different locations. As a result, we have 30 scenes under
varying lighting conditions in the dataset. Table \ref{gesture-table} has full
breakdown of the dataset, and Figure \ref{image-stats} has image statistics of
our dataset.


\section{Mobile Object Detection Model for Gesture Recognition}
\subsection{MobileNet SSD Architecture}
\begin{figure}[h]
  \centerline{\includegraphics[width=14cm]{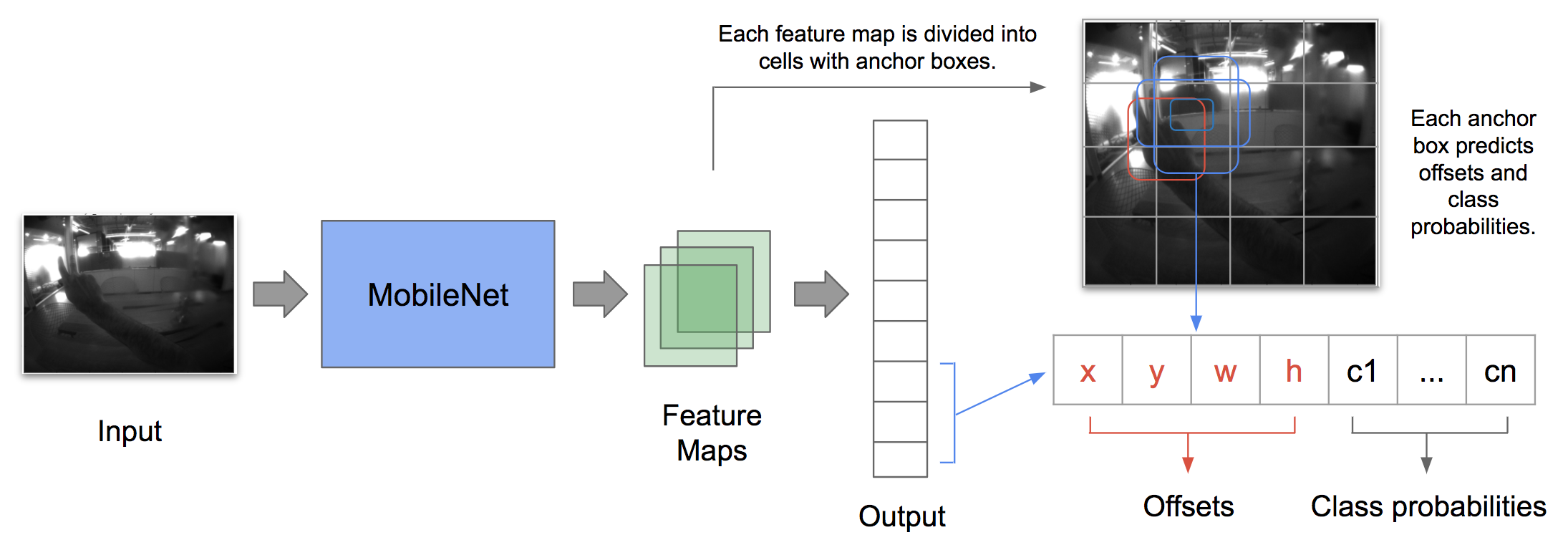}}
  \vspace{-5pt}
  \caption{Illustration of the MobileNet SSD model.}
  \label{ssd-figure}
  \vspace{-5pt}
\end{figure}

We trained a gesture recognition CNN based on TensorFlow
Object Detection API \cite{tf-object}. It is composed of two parts conceptually: a MobileNet \cite{mobilenet}
feature extractor to produce feature maps, and a SSD \cite{SSD}
multibox detector to predict bounding box location and gesture labels
(Figure \ref{ssd-figure}). For each anchor box in the SSD head, the model
predicts 4 offset values of the bouding box, and 9 class labels (4 distint gestures multiplexed with
either \verb+left+ or \verb+right+ hand, and one $\texttt{None}$ class). We use a cross entropy loss for classification, and smooth $\textit{L}_1$ loss as in
\cite{fast-rcnn} for bounding box localization loss. We add the two losses
together as the final loss function.

After model inference, we pick the bounding
box proposal with the highest confidence in label prediction, as we only expect
one label per image. SSD model
natively supports multi-class and multi-instance prediction too.

\subsection{Experiments and Results}
\label{section:augmentation}
Our models are trained with TensorFlow. Our training set contains 342,227 frames, and
evaluation set contains 64,354 frames. The same person only appears in one split.
For training, we use a batch size of 32. Image sizes are 320$\times$240, and of
one single color channel. We add random data augmentation to the training
dataset, including brightness and contrast perturbation, and random crop and
padding. 

On our evaluation dataset, we test the top confidence
bounding box prediction against the groundtruth label. Detailed results of the
model performance can be found in Table \ref{results-table}.

\subsection{Mobile Inference}
The trained TensorFlow models can be exported to run on mobile devices. We
benchmarked them on 
SnapDragon 821 chipset, which is common among Android devices since 2016. 
All results in Table \ref{results-table} reflect model inference time on
\textbf{one} Big CPU core on device.

\begin{table}[t]
  \centering
  \begin{tabular}{lllll}
    \toprule
    Model & Depth multiplier & Precision & \makecell[l]{Inference latency\\(ms)}
    & \makecell[l]{Total latency\\(ms)} \\
    \midrule
    MobileNetSSD-25\%  & 0.25 & 76.15\% &  31.8504 & 36.1658 \\ 
    MobileNetSSD-50\%  & 0.5 & 77.43\% &  77.4913 & 81.6922 \\ 
    MobileNetSSD-100\%  & 1 & 80.94\% &  265.2109 & 269.4694 \\ 
    \bottomrule
  \end{tabular}
  \vspace{-5pt}
  \caption{Results of model performance. Total latency
  includes inference, pre- and post-processing.}
  \vspace{-15pt}
  \label{results-table}
\end{table}

In our application, we chose the MobileNet-0.25 model. Our inference framerate
is at 30 frames per second on device. Accounting for pre- and post-processing
steps, the whole gesture detection pipeline can run at 27 fps sustainably.

\section{Conclusion}
In this work, we present a mobile egocentric gesture recognition pipeline. We built a mobile mixed-reality data capture tool, with which we can
automatically annotate gestures and bounding box locations. We created the largest-to-date egocentric gesture and
bounding boxes dataset. We trained a neural network based on the TensorFlow
Object Detection API \cite{tf-object}, and achieved 76.41\% precision and
real-time performance on mobile devices.

As future work, our \textit{label as you go} approach can be adapted to other data 
collection tasks, such as keypoint and segmentation mask annotation. It can also
be deployed on smartphones, where users could be asked to move their
phone such that the object in the viewfinder fits into the rendered target.

\bibliography{references}
\end{document}